\documentclass[11pt]{article}
\usepackage[preprint]{acl}
\usepackage{times}
\usepackage{latexsym}
\usepackage[T1]{fontenc}
\usepackage[utf8]{inputenc}
\usepackage{microtype}
\IfFileExists{inconsolata.sty}{\usepackage{inconsolata}}{}
\usepackage{graphicx}
\usepackage{amsmath,amssymb,amsthm}
\usepackage{booktabs}
\usepackage{xcolor}
\usepackage{listings}
\usepackage{enumitem}
\usepackage{tikz}
\usetikzlibrary{arrows.meta,positioning,calc,decorations.pathreplacing}

\definecolor{cblue}{HTML}{2A78D6}
\definecolor{corange}{HTML}{EB6834}
\definecolor{caqua}{HTML}{1BAF7A}
\definecolor{cink}{HTML}{3D3D3A}
\definecolor{cgray}{HTML}{8A8983}
\definecolor{cline}{HTML}{B5B4AD}
\tikzset{fig/.style={font=\scriptsize, line width=0.6pt, every node/.append style={text=cink}},
  card/.style={draw=#1, fill=#1!7, rounded corners=2pt, align=left, inner sep=3pt,
    execute at begin node={\hyphenpenalty=10000\exhyphenpenalty=10000}},
  arr/.style={-{Stealth[length=3.2pt]}, draw=cink, line width=0.6pt}}

\newif\ifplaceholder
\placeholderfalse

\newcommand{\AccNone}{45.5}
\newcommand{\AccName}{70.0}
\newcommand{\AccDec}{71.8}

\newcommand{\AccAll}{55.5}

\newcommand{\KeepOracle}{62.0}

\newcommand{\NDecisions}{110}

\newcommand{\NItems}{396}
\newcommand{\NNoise}{240}
\newcommand{\NDecisive}{56}
\newcommand{\NLate}{40}
\newcommand{\NBenign}{60}
\newcommand{\NSwitch}{60}

\newcommand{\NKeepRelated}{37}
\newcommand{\NKeepQuiet}{13}

\newcommand{\ItemsPerWindow}{115.7}

\newcommand{\LateWrongOracle}{94}

\newcommand{\BenignWrongOracle}{11}

\newcommand{\PAllDec}{0.003}

\newcommand{\PNameNone}{<0.001}

\newcommand{\VTDec}{75}
\newcommand{\VTSwitch}{48}

\newcommand{\VTKeepRelated}{27}
\newcommand{\VTKeepQuiet}{0}
\newcommand{\VTItems}{458}
\newcommand{\VTDecisive}{48}
\newcommand{\VTLate}{77}
\newcommand{\VTBenign}{108}
\newcommand{\VTNoise}{225}

\newcommand{\VTItemsPerWindow}{128.2}

\newcommand{\VOneBmRec}{97.1}

\newcommand{\VAccClBm}{97.3}

\newcommand{\VAccClBmTime}{97.3}

\newcommand{\VAccClOrRel}{96.0}

\newcommand{\VAccClAll}{96.0}

\newcommand{\VAccGptBm}{98.7}

\newcommand{\VAccGptBmTime}{98.7}

\newcommand{\VAccGptOrRel}{100.0}

\newcommand{\VAccGptAll}{100.0}

\newcommand{\VAccQwBm}{85.3}

\newcommand{\VAccQwBmTime}{92.0}

\newcommand{\VAccQwAll}{78.7}

\newcommand{\VAccNoneRange}{36.0}

\newcommand{\VAccNameRange}{72.0--73.3}

\newcommand{\VAccBmRange}{85.3--98.7}

\newcommand{\VPBmTimeBmQw}{=0.12}
\newcommand{\VPBmTimeBmQwWins}{6}
\newcommand{\VPBmTimeBmQwLosses}{1}

\newcommand{\VPPoolBmName}{<0.001}
\newcommand{\VPPoolBmNameWins}{53}
\newcommand{\VPPoolBmNameLosses}{6}

\newcommand{\VPPoolBmDecWins}{11}
\newcommand{\VPPoolBmDecLosses}{4}

\newcommand{\VPPoolBmWatchWins}{8}
\newcommand{\VPPoolBmWatchLosses}{5}

\newcommand{\VPPoolBmDepWins}{6}
\newcommand{\VPPoolBmDepLosses}{3}

\newcommand{\VPPoolOrDecOrRel}{=0.002}
\newcommand{\VPPoolOrDecOrRelWins}{10}

\newcommand{\VRobAccNone}{61.6}

\newcommand{\VRobAccBm}{89.0}

\newcommand{\VRobAccOrRel}{94.5}

\newcommand{\VTokRatioAllBm}{4.0}

\newcommand{\VRecBm}{100.0}

\newcommand{\VThItemsPerWindow}{78.8}

\newcommand{\VThRoleDecisive}{64}
\newcommand{\VThRoleLate}{48}
\newcommand{\VThRoleOver}{48}
\newcommand{\VThRoleHarmless}{48}
\newcommand{\VThRoleIrrelevant}{48}
\newcommand{\VThRoleDecoy}{128}
\newcommand{\VThRoleNoise}{128}

\newcommand{\VThN}{128}
\newcommand{\VThNKeep}{64}
\newcommand{\VThItems}{512}
\newcommand{\VThDossier}{12}

\newcommand{\WTrapBm}{17}

\newcommand{\WRecBmAlias}{37.5}
\newcommand{\WRecBmUp}{12.5}

\newcommand{\WTrapBmDoss}{89}

\newcommand{\WRecWatchBm}{92.2}

\newcommand{\WTrapWatchBm}{20}

\newcommand{\WTrapsOpen}{96}

\newcommand{\WAccClBm}{78.9}

\newcommand{\WAccClBmDoss}{93.0}

\newcommand{\WAccClWatchBm}{91.4}

\newcommand{\WKpClOrRel}{100.0}

\newcommand{\WKpClOrRelLate}{100.0}
\newcommand{\WKpClOrRelOver}{100.0}
\newcommand{\WKpClOrRelHarmless}{100.0}
\newcommand{\WKpClOrRelIrrel}{100.0}

\newcommand{\WAccClAll}{91.4}

\newcommand{\WKpQwOrRel}{40.6}

\newcommand{\WKpQwOrRelLate}{31.2}
\newcommand{\WKpQwOrRelOver}{0.0}
\newcommand{\WKpQwOrRelHarmless}{68.8}
\newcommand{\WKpQwOrRelIrrel}{62.5}

\newcommand{\WKpSmMinistralBOrRel}{18.8}

\newcommand{\WKpSmMinistralBOrRelLate}{0.0}
\newcommand{\WKpSmMinistralBOrRelOver}{0.0}
\newcommand{\WKpSmMinistralBOrRelHarmless}{50.0}
\newcommand{\WKpSmMinistralBOrRelIrrel}{25.0}

\newcommand{\WTokRatioAllDoss}{3.1}

\newcommand{\WPAllAllWatchBm}{=0.033}
\newcommand{\WPAllAllWatchBmW}{20}
\newcommand{\WPAllAllWatchBmL}{37}
\newcommand{\WPAllOrDecOrRel}{<0.001}
\newcommand{\WPAllOrDecOrRelW}{91}

\newcommand{\WPSmallWatchBmBmDoss}{=0.035}
\newcommand{\WPSmallWatchBmBmDossW}{12}
\newcommand{\WPSmallWatchBmBmDossL}{3}

\newcommand{\WPClBmDossBm}{<0.001}
\newcommand{\WPClBmDossBmW}{22}
\newcommand{\WPClBmDossBmL}{4}
\newcommand{\WPClWatchBmBm}{=0.002}
\newcommand{\WPClWatchBmBmW}{21}
\newcommand{\WPClWatchBmBmL}{5}
\newcommand{\WPClWatchBmBmDoss}{=0.77}
\newcommand{\WPClWatchBmBmDossW}{5}
\newcommand{\WPClWatchBmBmDossL}{7}

\newcommand{\WKOneBmDoss}{85.9}

\newcommand{\WKOneWatchBm}{71.9}

\newcommand{\orig}{\textsc{Direct}}
\newcommand{\stress}{\textsc{Stress}}
\newcommand{\indirect}{\textsc{Indirect}}
\newtheorem{proposition}{Proposition}
\title{Fresh at Decision, Stale at Action: Routing New Information to LLM Decisions Already Made}
\author{Ayoub Merimi \and Souhail Bakkali \and Abdeljalil Sakat}

\begin{document}
\maketitle

\begin{abstract}
Freshness shapes an LLM decision at the moment it is made; the action it commits to is latent, carried out later, in a changing world. We therefore introduce a timestamp theory of decision validity and a decision ledger that realises it. Six timestamps define a decision; a freshness ceiling bounds what up-to-date evidence guarantees at action time, placing the remaining risk in information published afterwards. Accuracy then factorises into two capabilities: routing, which delivers that information to affected decisions before their revision deadlines, and temporal judgment, which reads it for the time of action. The ledger is persistent state holding each decision's times, evidence and background knowledge, indexed by watch conditions written at decision time. Controlled decision streams with exact labels confirm four predictions across five LLMs, from 8B to frontier scale: freshness alone yields exactly the accuracy the ceiling predicts; models act on decisive news once it arrives, so accuracy follows routing; the ledger's index sets what routing reaches, and watch conditions reach indirect dependencies at low exposure; and the remaining errors concern when a change takes effect. Freshness makes a decision right when it is made; routing and temporal judgment keep it right when it is carried out.
\end{abstract}

\section{Introduction}
\label{sec:intro}

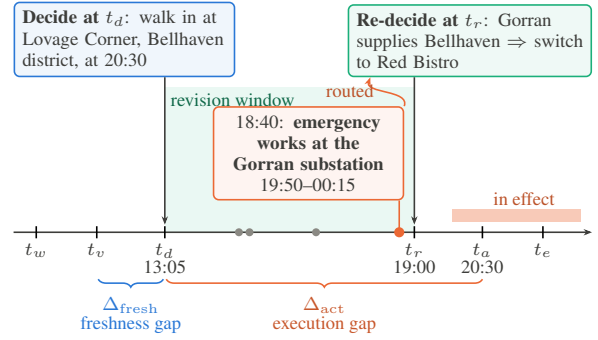
\begin{figure}[t]
\centering
\begin{tikzpicture}[fig, x=1cm, y=1cm]
  \fill[caqua!9] (2.0,0) rectangle (5.3,1.92);
  \node[anchor=north west, inner sep=2pt, text=caqua!60!black] at (2.0,1.92) {revision window};
  \draw[arr] (0,0) -- (7.65,0);
  \fill[corange!35] (5.8,0.14) rectangle (7.5,0.3);
  \node[anchor=south, inner sep=1.5pt, text=corange!80!black] at (6.75,0.3) {in effect};
  \foreach \x in {2.98, 3.12, 4.0} { \fill[cgray] (\x,0) circle (1.5pt); }
  \draw[corange] (5.1,0) -- (5.1,0.42);
  \fill[corange] (5.1,0) circle (1.9pt);
  \node[card=corange, anchor=south, text width=2.3cm, align=center] (news) at (3.9,0.42)
    {18:40: \textbf{emergency works at the Gorran substation} 19:50--00:15};
  \foreach \x/\t/\h in {0.3/w/{}, 1.1/v/{}, 2.0/d/13:05, 5.3/r/19:00, 6.2/a/20:30, 7.0/e/{}} {
    \draw (\x,0.08) -- (\x,-0.08);
    \node[anchor=north, inner sep=1pt, align=center] at (\x,-0.1) {$t_\t$\\[-1pt]\h};
  }
  \draw[decorate, decoration={brace, mirror, amplitude=3pt}, draw=cblue] (1.12,-0.62) -- (1.98,-0.62)
    node[midway, below=3pt, align=center, text=cblue] {$\Delta_{\mathrm{fresh}}$\\[-1pt]freshness gap};
  \draw[decorate, decoration={brace, mirror, amplitude=3pt}, draw=corange] (2.02,-0.62) -- (6.2,-0.62)
    node[midway, below=3pt, align=center, text=corange!85!black] {$\Delta_{\mathrm{act}}$\\[-1pt]execution gap};
  \node[card=cblue, anchor=south west, text width=2.7cm] (dec) at (0,2.05)
    {\textbf{Decide at $t_d$}: walk in at Lovage Corner, Bellhaven district, at 20:30};
  \node[card=caqua, anchor=south east, text width=3.0cm] (re) at (7.65,2.05)
    {\textbf{Re-decide at $t_r$}: Gorran supplies Bellhaven $\Rightarrow$ switch to Red Bistro};
  \draw[arr] (2.0,2.05) -- (2.0,0.1);
  \draw[arr] (5.3,2.05) -- (5.3,0.1);
  \draw[arr, draw=corange] (news.north east) to[out=70, in=250] node[left, pos=0.55, text=corange!80!black] {routed} ($(re.south west)+(0.3,0)$);
\end{tikzpicture}
\caption{A decision from the \indirect{} stream on its six timestamps. Retrieval at decision time shrinks the freshness gap $\Delta_{\mathrm{fresh}}$. The notice that settles the decision is published inside the execution gap $\Delta_{\mathrm{act}}$, is in effect at action time, and names only the substation that supplies the restaurant's district; background knowledge in the ledger routes it to the decision before the revision deadline. Grey dots are other notices in the window.}
\label{fig:timeline}
\end{figure}

At 13:05, a user asks an assistant for a walk-in dinner at 20:30. The assistant chooses Lovage Corner, the best-rated restaurant within walking distance, in the Bellhaven district. At 18:40, the grid operator announces emergency works at the Gorran substation from 19:50, leaving every area it supplies without power; the Gorran substation supplies the Bellhaven district (Figure~\ref{fig:timeline}). The choice was right at 13:05 and is wrong at 20:30. The information that settles it appeared after the decision and refers to the restaurant only through the substation behind its district. This decision comes from our test streams (Appendix~\ref{app:example}).

The example separates two properties of a decision. \emph{Evidence freshness} describes what was known when the choice was made. \emph{Action validity} describes whether the choice is right when it is carried out. Research on time in language models secures the first: benchmarks track how models age \citep{lazaridou2021mind, dhingra2022time, kim2024carpe}, test answers against evolving news \citep{liska2022streamingqa, kasai2023realtime}, and refresh models with search \citep{vu2024freshllms}. This framing fits question answering, where an answer is used the moment it is produced. LLM agents produce decisions that are carried out later, such as itineraries \citep{xie2024travelplanner} and service actions \citep{yao2025taubench}, and these are judged by the second.

This paper develops a theory of action validity and answers the title question: freshness is necessary and insufficient. Six timestamps define a decision, from when its evidence is written to when its outcome is judged. A \emph{freshness ceiling} shows that, under a constant rate of change, the probability that evidence still holds when the user acts is bounded by the execution gap between deciding and acting, whatever the freshness at decision time. The remaining risk lies in information published inside that gap. Decision accuracy then factorises into \emph{routing}, whether that information reaches the decision before its revision deadline, and \emph{judgment}, whether the model reads it correctly for the time of action (Section~\ref{sec:theory}).

The theory comes with a model of the system that realises it. A \emph{decision ledger} is persistent state \citep{bakkali2026amnesia} that stores each decision with its times, its evidence and its background knowledge, and a router delivers each new item to the open decisions it matches (Section~\ref{sec:ledger}). A query retrieves what the user asks about today; a ledger retrieves what yesterday's decisions still need. The ledger also makes the theory testable: what a router can reach follows from what the ledger indexes, and what the model does with a delivered item follows from its judgment.

We derive four predictions and test them in controlled decision streams, in which every label follows from timestamps and stated facts, under three conditions that vary how news refers to a decision and how hard its negatives are: decisive news that names the option or its stated dependency (\orig{}), the same with hard negatives (\stress{}), and news that reaches the option through other names and upstream entities (\indirect{}). The tests confirm all four predictions across five LLMs, from Ministral 8B to GPT-6.1 Sol.
\begin{itemize}
  \item \textbf{P1, the freshness ceiling holds.} With perfect evidence at decision time and nothing afterwards, accuracy equals the share of decisions that should stand.
  \item \textbf{P2, delivery is sufficient for change.} Once decisive news arrives, models switch in 91--100\% of cases under every router, so the accuracy of changing decisions tracks routing recall, and recall follows the ledger's index: the decision text reaches \WRecBmAlias\% of news that uses another name for a dependency, and background knowledge in the ledger reaches all of it.
  \item \textbf{P3, misleading deliveries cost correct decisions.} Every misleading item delivered is a chance to abandon a decision that should stand. Watch conditions written at decision time reach \WRecWatchBm\% of decisive news with \WTrapWatchBm{} misleading deliveries against \WTrapBmDoss{} for the whole background, and broadcasting helps only models that judge well.
  \item \textbf{P4, the residual error is temporal judgment.} Given exactly the related news, Ministral 8B and Qwen3-235B abandon every decision whose related change ended before the activity, Qwen3-8B abandons \LateWrongOracle\% of those whose change starts after it, and Claude Sonnet 4.6 keeps all of them.
\end{itemize}

This is a theory paper with controlled empirical tests. It contributes a temporal theory of LLM decisions with a freshness ceiling and a routing--judgment factorisation (Section~\ref{sec:theory}), the decision ledger as a model of the system that realises it, including watch conditions (Section~\ref{sec:ledger}), and controlled tests that confirm the theory's predictions across five models (Sections~\ref{sec:streams}--\ref{sec:results}).

\section{Related Work}
\label{sec:related}

\paragraph{Freshness and temporal knowledge.}
Language models degrade on text written after their training data \citep{lazaridou2021mind} and can be probed and refreshed for time-dependent facts \citep{dhingra2022time, kim2024carpe}. Benchmarks over evolving news and time-sensitive questions test whether answers are current \citep{chen2021timeqa, liska2022streamingqa, kasai2023realtime}, search augmentation keeps them current \citep{lewis2020rag, vu2024freshllms}, and evaluation sets themselves go stale as facts change \citep{jiang2026benchmarksage}. Temporal reasoning benchmarks test ordering and arithmetic over time \citep{fatemi2025testoftime}. These works judge the model when it answers; we judge a decision when it is carried out.

\paragraph{Updates, conflicts and revision.}
Knowledge editing changes stored facts \citep{meng2022rome}, and multi-hop and ripple-effect evaluations require an update to reach every fact that depends on it \citep{zhong2023mquake, cohen2024ripple}. Models resolve conflicts between parametric and contextual knowledge unevenly \citep{xie2024adaptive, xu2024conflicts}, including conflicts that arise while agents act \citep{lyu2026kcbench}, and revise conclusions under new premises with mixed success \citep{kassner2021beliefbank, wilie2024belief}. Post-decision routing carries the ripple-effect requirement to decisions: a new fact must reach every open decision that depends on it, and the model must then revise it correctly.

\paragraph{Agents that act over time.}
Agent benchmarks evaluate plans and actions executed after reasoning \citep{yao2023react, xie2024travelplanner, yao2025taubench}. Time-aware simulations score how agents schedule work \citep{zhang2024timearena}, and agents misjudge when elapsed time calls for fresh tool calls \citep{cheng2026temporallyblind}. Proactive agents decide when to intervene \citep{lu2025proactive, tang2026proagentbench, li2026proevent, vibelife2026}, maintain long-term intents through trigger conditions \citep{shi2026longterm}, and accept user corrections during execution \citep{zhai2026revisable}. We study many concurrent decisions with revision deadlines, a shared stream of world news, and the cost of deciding which news reaches which decision.

\paragraph{Memory and retrieval.}
Agent memories store and retrieve past interactions \citep{park2023generative, packer2023memgpt, chhikara2025mem0, xu2025amem}, invalidate outdated facts in temporal graphs \citep{rasmussen2025zep}, index memories by time and topic \citep{tian2026swiftmem}, and are evaluated on long-term recall \citep{wu2025longmemeval, du2026memory}. Persistent, evidence-grounded state with temporal validity is a design requirement for systems that work across sessions \citep{bakkali2026amnesia}; the decision ledger applies it to commitments. Watch conditions are document expansion \citep{nogueira2019doc2query} and generated queries \citep{gao2023hyde} written once, at decision time. BM25 remains a strong zero-shot retriever next to dense retrieval \citep{karpukhin2020dpr, thakur2021beir}, and long contexts degrade reasoning when relevant content is buried or lacks literal overlap \citep{shi2023distracted, liu2024lost, modarressi2025nolima}.

\section{A Theory of Decisions Over Time}
\label{sec:theory}

\paragraph{Six timestamps.}
Evidence is written at $t_w$ and last verified at $t_v$. The model decides at $t_d$ with evidence verified at $t_v \le t_d$. The decision can be revised until the revision deadline $t_r$, is carried out at $t_a$, and its outcome is observed at $t_e$, with $t_d < t_r \le t_a \le t_e$. The revision deadline captures commitment: an order ships, a table is taken, a train departs.

\paragraph{The freshness ceiling.}
Let $s(t)$ be the state of the world and $U(a, s)$ the utility of action $a$ in state $s$. The model chooses $\hat a$ at $t_d$, and the outcome is $U(\hat a, s(t_a))$, so the target is the best action at $t_a$. The age of the evidence at action time decomposes as
\begin{equation}
  t_a - t_v = \underbrace{(t_d - t_v)}_{\Delta_{\mathrm{fresh}}} + \underbrace{(t_a - t_d)}_{\Delta_{\mathrm{act}}}.
\end{equation}
Retrieval and knowledge updates shrink the freshness gap $\Delta_{\mathrm{fresh}}$; the execution gap $\Delta_{\mathrm{act}}$ is set by the task. If a fact $f$ changes at a constant hazard rate $\lambda_f$, it still holds $\tau$ time units after verification with probability $e^{-\lambda_f \tau}$, so
\begin{equation}
  \Pr[f \text{ holds at } t_a] = e^{-\lambda_f \Delta_{\mathrm{fresh}}}\, e^{-\lambda_f \Delta_{\mathrm{act}}}.
  \label{eq:ceiling}
\end{equation}
\begin{proposition}[Freshness ceiling]
\label{prop:ceiling}
For any fact $f$ and execution gap $\Delta_{\mathrm{act}}$, the probability that $f$ holds at action time is at most $e^{-\lambda_f \Delta_{\mathrm{act}}}$, for every degree of freshness.
\end{proposition}
Appendix~\ref{app:proofs} proves the proposition and extends it to decisions with several dependencies. Freshness is a property of the evidence; validity at action time is a property of the whole pipeline. The ceiling locates the residual risk in the execution gap and in the volatility of the evidence, and only information that arrives after $t_d$ can reduce it: a notice published inside the execution gap that reaches the decision before $t_r$.

\paragraph{Post-decision routing.}
A ledger holds open decisions. Decision $i$ has a chosen option $a_i$, alternatives, evidence $E_i$, times $t_d^i < t_r^i \le t_a^i$ and an activity $[t_a^i, t_a^i + h_i]$. A stream delivers items $x_j$, each written and published at $t_w^j$. Item $j$ is \emph{decisive} for decision $i$ if it is published in the revision window, $t_d^i < t_w^j < t_r^i$, concerns $a_i$ or an entity $a_i$ depends on, and blocks $a_i$ over an interval $I_j$ that overlaps the activity, so that $a_i$ stops being the best option. An item about $a_i$ or its dependencies that is not decisive is \emph{related}: a change outside the activity, or news that leaves $a_i$ usable. A router assigns each item to a subset of the decisions open at $t_w^j$, and at $t_r^i$ the model reads $E_i$ with the items routed to $i$ and keeps or replaces $a_i$.

\paragraph{Routing and judgment.}
Let $\pi$ be the share of decisions that should change. For those, let $\rho$ be the routing recall, the probability that a decisive item reaches the decision; $S$ the probability that the model switches when it does; and $L$ the probability that it switches when it does not. For decisions that should stand, let $K$ be the probability that the model keeps them. Final accuracy is
\begin{equation}
  A = (1-\pi)\,K + \pi\,\big[\rho S + (1-\rho) L\big].
  \label{eq:decomp}
\end{equation}
Routing sets $\rho$ and the items each decision sees; judgment sets $S$, $L$ and $K$ (Appendix~\ref{app:proofs} derives Eq.~\ref{eq:decomp} and the trade-off between routers it implies). A routing failure leaves a decision stale; a judgment failure ignores decisive news or abandons a decision that should stand. A decisive item reaches the chosen option in one of four ways: it names the option (Lovage Corner), the dependency stated in the evidence (the Bellhaven district), another name for that dependency (Chime Green, as the district is known locally), or an entity upstream of it (the Gorran substation, which supplies the district). Each step down this list moves the link from the decision's text into background knowledge.

\paragraph{Predictions.}
The theory makes four predictions.
\textbf{P1:} without post-decision routing ($\rho = 0$, $L = 0$, $K = 1$), accuracy equals $1-\pi$, the share of decisions that should stand, for every model.
\textbf{P2:} delivery is sufficient for change ($S \approx 1$), so the accuracy of changing decisions tracks $\rho$, and $\rho$ is set by whether the router's index contains the entity a decisive item names.
\textbf{P3:} each misleading item delivered lowers $K$ at a rate set by the model's judgment, so routers trade recall against exposure, and broadcasting, which maximises both, helps only models that judge well.
\textbf{P4:} with oracle routing ($\rho = 1$, every related item and nothing else), the residual error concentrates on temporal judgment: deciding whether $I_j$ overlaps the activity.

\section{The Decision Ledger}
\label{sec:ledger}

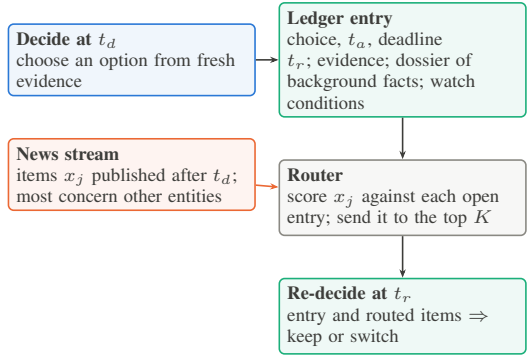
\begin{figure}[t]
\centering
\begin{tikzpicture}[fig, node distance=0.35cm and 0.3cm, every node/.append style={text width=3.05cm}]
  \node[card=cblue] (dec) {\textbf{Decide at $t_d$}\\choose an option from fresh evidence};
  \node[card=caqua, right=of dec] (led) {\textbf{Ledger entry}\\choice, $t_a$, deadline $t_r$; evidence; dossier of background facts; watch conditions};
  \node[card=corange, below=0.55cm of dec] (str) {\textbf{News stream}\\items $x_j$ published after $t_d$; most concern other entities};
  \node[card=cgray, below=0.55cm of led] (rout) {\textbf{Router}\\score $x_j$ against each open entry; send it to the top $K$};
  \node[card=caqua, below=0.55cm of rout] (re) {\textbf{Re-decide at $t_r$}\\entry and routed items $\Rightarrow$ keep or switch};
  \draw[arr] (dec) -- (led);
  \draw[arr] (led) -- (rout);
  \draw[arr, draw=corange] (str) -- (rout);
  \draw[arr] (rout) -- (re);
\end{tikzpicture}
\caption{The decision ledger. Routing uses no model calls; the model is called again only for decisions that receive news. What the entry indexes sets what the router can reach.}
\label{fig:pipeline}
\end{figure}

Each ledger entry is persistent state that outlives the session in which the decision was made \citep{bakkali2026amnesia}. When the model decides at $t_d$, it writes the entry: the chosen option and its alternatives, the action time $t_a$, the revision deadline $t_r$, the evidence about the chosen option, and a \emph{dossier} of background facts, such as other names of the entities the option depends on and what those entities rely on in turn. In deployment, the dossier holds what the assistant looked up while deciding, including much that is irrelevant to the user's activity. When an item $x_j$ is published, a router scores the open entries and sends the item to the $K$ highest-scoring ones. At the revision deadline, the model reads the entry and the routed items in order of publication and keeps or switches. Decisions that receive no items keep their choice without a model call. Routers differ in what they index, which by P2 sets what they can reach.

\paragraph{Decision text.} BM25 or dense embeddings compare the item with the decision's request and its evidence about the chosen option, reaching news that names the option or its stated dependency.

\paragraph{Whole dossier.} The decision text is expanded with every dossier fact, reaching news about other names and upstream entities together with news about everything else the dossier mentions.

\paragraph{Watch conditions.} At $t_d$, the model reads the entry and writes three to eight conditions $w_{i1}, \dots, w_{ik}$ describing events that would stop the user's activity. The conditions name every entity the activity depends on, including other names and upstream entities, and leave out what the user does not need (Appendix~\ref{app:prompts}). The router indexes the decision text with the conditions (BM25) or scores each decision by its best condition, $\max_k \cos(e(x_j), e(w_{ik}))$, with a sentence encoder $e$. Watch conditions are document expansion \citep{nogueira2019doc2query} performed once, at write time, by the component that knows why the option was chosen; by P3 they should match the dossier's reach with less exposure.

\paragraph{Time filter.} A rule drops an item for a decision when the item states a start time after the decision's activity ends, performing by rule the temporal check of P4.

\section{Controlled Decision Streams}
\label{sec:streams}

Testing the predictions requires exact control over which items are decisive, how they refer to the option, and when changes take effect. Real logs offer none of this, and real entities let models answer from memory. We therefore generate fictional decisions and news streams in which every label follows from timestamps and stated facts; the generator and data are released with the paper.

\paragraph{Decisions and labels.}
Each decision has two options. Option~1 is better at $t_d$ (higher rating, lower price, faster journey, better views) and is the model's initial choice; option~2 is acceptable but worse. Option~1 depends on an entity named in its evidence: a district, a rail operator, a sales or permit platform, a mountain range or a shopping centre. Lags between deciding and acting range from three hours to three days, and revision deadlines precede the action by 0.5 to 2 hours. A decision's gold answer is option~2 if some item published in its revision window targets option~1, its dependency or an entity upstream of it and blocks option~1 during the activity, and option~1 otherwise. The generator excludes decisions invalidated before they were made and asserts every label (Appendix~\ref{app:bench}).

\paragraph{Three conditions.}
Each condition stresses a different part of the theory. \orig{} measures every term of Eq.~\ref{eq:decomp} with an open model; \stress{} adds hard negatives that name the option, testing exposure (P3) and temporal judgment (P4); \indirect{} moves the reference from the decision text into background knowledge, testing whether the index sets recall (P2). \orig{} spans five domains with independent random news per decision: a decisive item ($p{=}0.5$) that names the option or its stated dependency, a late item ($p{=}0.3$) that blocks the option only after the activity, a harmless item ($p{=}0.5$), and two unrelated items. \stress{} adds, for every decision, a \emph{same-name decoy}, a blocking notice about a different entity whose name extends the option's (``Clove Fork Annex''), and, each with probability 0.85, a late and a harmless notice naming the option itself; eleven rules paraphrase event wording. \indirect{} spans four domains and fixes every proportion. Half of the decisions should switch, and their decisive notice uses each of the four ways of reference for 16 decisions; only the first two match the decision text. The other half should stand and receive one of four hard negatives, each for 16 decisions: a change that starts after the activity, a change that is over before it starts, harmless news, and a disruption to a facility the user does not need (a tram line for a user who walks). Every decision receives exactly two related notices and one decoy, so neither the amount nor the wording of the news reveals the label. Each \indirect{} dossier holds \VThDossier{} facts: the other name, the upstream entity, four unneeded facilities, and six \emph{traps} that mention entities belonging to other decisions in irrelevant roles (``Tram line T12 draws its power from the Hensall traction substation''). Table~\ref{tab:stats} summarises the conditions.

\begin{table}[t]
\centering
\small
\setlength{\tabcolsep}{3pt}
\resizebox{\columnwidth}{!}{%
\begin{tabular}{l rrr}
\toprule
 & \orig{} & \stress{} & \indirect{} \\
\midrule
Decisions & \NDecisions{} & \VTDec{} & \VThN{} \\
\quad should switch & \NSwitch{} & \VTSwitch{} & \VThRoleDecisive{} \\
\quad\quad via name or stated dependency & \NSwitch{} & \VTSwitch{} & 32 \\
\quad\quad via other name or upstream & 0 & 0 & 32 \\
\quad should stand & \the\numexpr\NKeepRelated+\NKeepQuiet\relax{} & \the\numexpr\VTKeepRelated+\VTKeepQuiet\relax{} & \VThNKeep{} \\
News items & \NItems{} & \VTItems{} & \VThItems{} \\
\quad decisive & \NDecisive{} & \VTDecisive{} & \VThRoleDecisive{} \\
\quad related, not decisive & \the\numexpr\NLate+\NBenign\relax{} & \the\numexpr\VTLate+\VTBenign\relax{} & \the\numexpr\VThRoleLate+\VThRoleOver+\VThRoleHarmless+\VThRoleIrrelevant\relax{} \\
\quad decoys and unrelated & \NNoise{} & \VTNoise{} & \the\numexpr\VThRoleDecoy+\VThRoleNoise\relax{} \\
Dossier facts per decision & -- & -- & \VThDossier{} \\
Items in a decision's window & \ItemsPerWindow{} & \VTItemsPerWindow{} & \VThItemsPerWindow{} \\
\bottomrule
\end{tabular}}
\caption{The three conditions. A decision's revision window holds dozens to hundreds of items, of which a handful concern it.}
\label{tab:stats}
\end{table}

\section{Experimental Setup}
\label{sec:setup}

\paragraph{Routers.}
All routers of a condition share one re-decision prompt and differ in which items reach each decision. \emph{None} routes nothing and realises perfect freshness at $t_d$. \emph{All items} broadcasts everything published while the decision is open. \emph{Name match} routes items that contain the option's name. The routers of Section~\ref{sec:ledger} send each item to its $K{=}3$ best-scoring open decisions; dense routing uses BGE-base \citep{xiao2024cpack}. \emph{Oracle: related} delivers exactly the items about the option, its dependencies and its facilities, including every hard negative, and \emph{Oracle: decisive} delivers only decisive items; both use gold annotations and isolate judgment, and \emph{Oracle: decisive} also signals that a change is due, so it serves as a ceiling. \stress{} also tests the time filter and \emph{dependency queries}, three fixed queries naming the option and its stated dependency.

\paragraph{Models.}
On \orig{}, Qwen3-8B with thinking disabled \citep{yang2025qwen3} runs locally with greedy decoding \citep{wolf2020transformers} and writes its own watch conditions. On \stress{}, we evaluate Claude Sonnet 4.6, GPT-6.1 Sol and Qwen3-235B-A22B; on \indirect{}, Claude Sonnet 4.6, Qwen3-235B-A22B and Ministral 8B, with Mistral 7B \citep{jiang2023mistral} reported in Appendix~\ref{app:more}. Hosted models run at temperature 0 with structured output, Claude writes the watch conditions for every model of a condition so that routing is identical across models, and every reported cell is complete with no imputed answers. Prompts and decoding settings are in Appendices~\ref{app:prompts} and~\ref{app:impl}.

\paragraph{Measures.}
We measure accuracy $A$ and the terms of Eq.~\ref{eq:decomp}: recall $\rho$, the switch rates $S$ and $L$, and the keep rate $K$. \emph{Traps} count items delivered to a decision whose dossier mentions their entity in an irrelevant role. Cost is input tokens, measured with Claude because prompts are identical across models. Exact McNemar tests compare routers on per-decision correctness, per model and pooled over the models of a condition; W--L counts decisions where the first router is right and the second wrong, and the reverse.

\section{Testing the Predictions}
\label{sec:results}

\IfFileExists{generated/v2_table.tex}{\begin{table}[t]
\centering
\small
\setlength{\tabcolsep}{3.2pt}
\resizebox{\columnwidth}{!}{%
\begin{tabular}{l rrrrrr}
\toprule
& \multicolumn{3}{c}{Routing} & \multicolumn{3}{c}{Accuracy (\%)} \\
\cmidrule(lr){2-4}\cmidrule(lr){5-7}
Router & Items & Rec. & Tok. & Claude & GPT & Qwen \\
\midrule
None & 0.0 & -- & 0.0 & 36.0 & 36.0 & 36.0 \\
Name match & 3.5 & 56 & 1.2 & 73.3 & 72.0 & 73.3 \\
BM25: decision & 18.0 & 100 & 2.0 & \textbf{97.3} & 98.7 & 85.3 \\
BM25: + time filter & 14.3 & 100 & 1.8 & \textbf{97.3} & 98.7 & \textbf{92.0} \\
Dense: decision & 18.0 & 96 & 2.0 & 90.7 & 96.0 & 85.3 \\
Dense: watch cond. & 18.0 & 96 & 2.0 & 93.3 & 97.3 & 86.7 \\
Dependency queries & 18.0 & 100 & 2.0 & 93.3 & 97.3 & 86.7 \\
\midrule
Oracle: related & 3.3 & 100 & 1.2 & 96.0 & 100.0 & 90.7 \\
Oracle: decisive & 0.6 & 100 & 0.7 & 100.0 & 100.0 & 100.0 \\
All items & 128.2 & 100 & 8.1 & 96.0 & \textbf{100.0} & 78.7 \\
\bottomrule
\end{tabular}}
\caption{\textsc{Stress}: final-decision accuracy (\%) of Claude Sonnet 4.6, GPT-6.1 Sol and Qwen3-235B on the same 75 decisions. Items: new items delivered per decision; Rec.: share of decisive item--decision pairs delivered; Tok.: Claude Sonnet 4.6 input tokens per decision for the re-decision calls (thousands). Bold: best router without gold labels.}
\label{tab:v2}
\end{table}
}{}
\IfFileExists{generated/v3_table.tex}{\begin{table}[t]
\centering
\small
\setlength{\tabcolsep}{3.5pt}
\resizebox{\columnwidth}{!}{%
\begin{tabular}{l rr rrr}
\toprule
& \multicolumn{2}{c}{Routing} & \multicolumn{3}{c}{Accuracy (\%)} \\
\cmidrule(lr){2-3}\cmidrule(lr){4-6}
Router & Recall & Traps & Claude & Qwen-235B & Ministral 8B \\
\midrule
None & -- & 0 & 50.0 & 50.0 & 50.0 \\
Name match & 25 & 0 & 62.5 & 57.0 & 52.3 \\
BM25: decision & 62 & 17 & 78.9 & 60.2 & 55.5 \\
BM25: + dossier & 100 & 89 & \textbf{93.0} & 61.7 & 54.7 \\
Dense: + dossier & 86 & 59 & 87.5 & 60.9 & 57.8 \\
BM25: + watch cond. & 92 & 20 & 91.4 & \textbf{64.1} & \textbf{61.7} \\
Dense: watch cond. & 84 & 27 & 85.9 & 57.8 & 58.6 \\
\midrule
Oracle: related & 100 & 0 & 99.2 & 70.3 & 59.4 \\
Oracle: decisive & 100 & 0 & 100.0 & 100.0 & 100.0 \\
All items & 100 & 96 & 91.4 & 56.2 & 56.2 \\
\bottomrule
\end{tabular}}
\caption{\textsc{Indirect}: final-decision accuracy (\%) on 128 balanced decisions. Recall: decisive notices delivered to their decision (\%); Traps: notices delivered to a decision whose dossier mentions their entity in an irrelevant role (of 96). Claude writes the watch conditions for every model. Bold: best router without gold labels.}
\label{tab:v3}
\end{table}
}{}

Table~\ref{tab:v2} reports \stress{}, Table~\ref{tab:v3} reports \indirect{}, and Appendix~\ref{app:orig} reports \orig{}.

\paragraph{P1: the freshness ceiling holds.}
\emph{None} has perfect evidence at decision time and nothing afterwards. Its accuracy equals the share of decisions that should stand for every model: \AccNone\% on \orig{}, \VAccNoneRange\% on \stress{} and 50.0\% on \indirect{}. Every decision that should change stays wrong, because the information that changes it was published after the decision. Routing any new information helps: name matching alone lifts Qwen3-8B on \orig{} to \AccName\% ($p\,\PNameNone$).

\begin{figure}[t]
\centering
\includegraphics{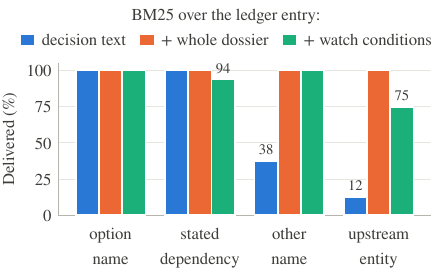}
\caption{Decisive notices delivered on \indirect{}, by how the notice refers to the chosen option (16 decisions each; unlabelled bars reach 100\%). The decision text reaches the option's name and its stated dependency; the dossier and watch conditions extend the reach to other names and upstream entities.}
\label{fig:reftype}
\end{figure}

\begin{figure}[t]
\centering
\includegraphics[width=\columnwidth]{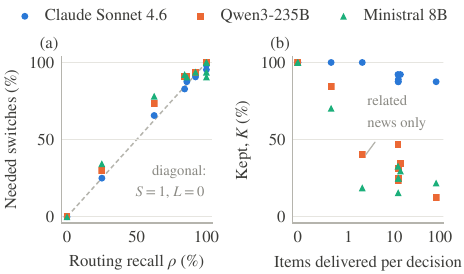}
\caption{The two factors of Eq.~\ref{eq:decomp} on \indirect{}, one point per router. (a) Needed switches against routing recall: on the diagonal, the model switches exactly when decisive news arrives. (b) Decisions kept among those that should stand, against the items delivered to them; the points at two items are the related-news oracle.}
\label{fig:factors}
\end{figure}

\paragraph{P2: delivery is sufficient for change, and the index sets recall.}
On \indirect{}, once a decisive notice reaches a decision, models switch in 90.6--100\% of cases, for every router and every model. Needed switches therefore follow routing recall (Figure~\ref{fig:factors}a): Claude's points lie on the diagonal, within 4.7 points of recall under every router, and wherever recall is below 100\% the weaker models lie above it, because they also switch on decisions whose decisive notice never arrived. Accuracy on changing decisions is a routing quantity, and recall follows the index (Figure~\ref{fig:reftype}). Name matching reaches only news that names the option. The decision text reaches every notice that names the option or its stated dependency, \WRecBmAlias\% of those that use another name, and \WRecBmUp\% of those that name an upstream entity, so Claude reaches \WAccClBm\%. Adding the dossier to the index reaches every decisive notice and lifts Claude to \WAccClBmDoss\% (\WPClBmDossBmW--\WPClBmDossBmL, $p\WPClBmDossBm$), and watch conditions lift it to \WAccClWatchBm\% (\WPClWatchBmBmW--\WPClWatchBmBmL, $p\WPClWatchBmBm$). When news names the stated dependency, as on \stress{}, the decision text is a sufficient index: BM25 delivers \VRecBm\% of decisive items and lifts the three models from \VAccNameRange\% with name matching to \VAccBmRange\% (\VPPoolBmNameWins--\VPPoolBmNameLosses{} pooled, $p\VPPoolBmName$), as accurate as oracle routing of all related news for Claude (\VAccClBm\% and \VAccClOrRel\%) and GPT (\VAccGptBm\% and \VAccGptOrRel\%). Watch conditions, dense similarity and idealised dependency queries route as many items there, and pooled over models BM25 wins more discordant decisions against each (\VPPoolBmWatchWins--\VPPoolBmWatchLosses, \VPPoolBmDecWins--\VPPoolBmDecLosses{} and \VPPoolBmDepWins--\VPPoolBmDepLosses).

\paragraph{P3: misleading deliveries cost correct decisions.}
Every item delivered to a decision that should stand is misleading, and Figure~\ref{fig:factors}b shows its cost: Claude keeps at least 87.5\% of these decisions under every router, while Qwen3-235B and Ministral 8B keep fewer than half under every BM25 and dense router. Indexing the whole dossier delivers \WTrapBmDoss{} of the \WTrapsOpen{} trap notices, because every fact it holds becomes a match. Watch conditions deliver \WRecWatchBm\% of the decisive notices and \WTrapWatchBm{} traps, close to the \WTrapBm{} of the decision text alone, because the model writes conditions only about what the activity needs. Under the retrieval and broadcast routers, a delivered trap raises the rate of needless switches for every model: from 9.4\% to 14.0\% for Claude, from 69.7\% to 74.8\% for Qwen3-235B and from 71.5\% to 84.1\% for Ministral 8B. Under retrieval routing, the weaker models also switch on 20--67\% of the changing decisions whose decisive notice never arrived, prompted by other news, so their gains from recall come with losses on $K$. Lower exposure therefore matters most where judgment is weakest: watch conditions give Qwen3-235B and Ministral 8B their best accuracy and beat the whole dossier for Ministral 8B (\WPSmallWatchBmBmDossW--\WPSmallWatchBmBmDossL, $p\WPSmallWatchBmBmDoss$), while for Claude the two indexes are equal (\WPClWatchBmBmDossW--\WPClWatchBmBmDossL, $p\WPClWatchBmBmDoss$). Broadcasting maximises both recall and exposure, at \VTokRatioAllBm{} times the tokens of BM25 routing on \stress{} and \WTokRatioAllDoss{} times on \indirect{}. Models that judge well absorb it (GPT \VAccGptAll\% and Claude \VAccClAll\% on \stress{}, Claude \WAccClAll\% on \indirect{}); the others lose: Qwen3-8B falls from \AccDec\% to \AccAll\% on \orig{} ($p=\PAllDec$), Qwen3-235B from \VAccQwBm\% to \VAccQwAll\% on \stress{}, and on \indirect{} broadcasting loses to watch conditions (\WPAllAllWatchBmW--\WPAllAllWatchBmL{} pooled, $p\WPAllAllWatchBm$).

\begin{table}[t]
\centering
\small
\setlength{\tabcolsep}{3.5pt}
\resizebox{\columnwidth}{!}{%
\begin{tabular}{l rrrr}
\toprule
 & \orig{} & \multicolumn{3}{c}{\indirect{}} \\
\cmidrule(lr){2-2}\cmidrule(lr){3-5}
Related news & Qwen3-8B & Ministral & Qwen-235B & Claude \\
\midrule
starts after the activity & \the\numexpr100-\LateWrongOracle\relax{} & \WKpSmMinistralBOrRelLate{} & \WKpQwOrRelLate{} & \WKpClOrRelLate{} \\
over before it starts & -- & \WKpSmMinistralBOrRelOver{} & \WKpQwOrRelOver{} & \WKpClOrRelOver{} \\
harmless & \the\numexpr100-\BenignWrongOracle\relax{} & \WKpSmMinistralBOrRelHarmless{} & \WKpQwOrRelHarmless{} & \WKpClOrRelHarmless{} \\
facility not needed & -- & \WKpSmMinistralBOrRelIrrel{} & \WKpQwOrRelIrrel{} & \WKpClOrRelIrrel{} \\
\midrule
all that should stand & \KeepOracle{} & \WKpSmMinistralBOrRel{} & \WKpQwOrRel{} & \WKpClOrRel{} \\
\bottomrule
\end{tabular}}
\caption{Judgment in isolation (P4): decisions kept (\%) when the oracle delivers exactly the related news and none of it changes the best option. On \indirect{}, each row except the last covers 16 decisions per model.}
\label{tab:keep}
\end{table}

\paragraph{P4: the residual error is temporal judgment.}
The oracles remove routing error (Table~\ref{tab:keep}). Given only decisive items, every model on \stress{} and \indirect{} is correct on every decision. Adding the related news that leaves the decision unchanged costs \WPAllOrDecOrRelW{} decisions across the three \indirect{} models and gains none ($p\WPAllOrDecOrRel$), and \VPPoolOrDecOrRelWins{} decisions to none on \stress{} ($p\VPPoolOrDecOrRel$). The errors concentrate on time. Ministral 8B and Qwen3-235B abandon every decision whose related change ended before the activity starts and keep \WKpSmMinistralBOrRelLate\% and \WKpQwOrRelLate\% of those whose change starts after it, more often than they abandon decisions for harmless news or an unneeded facility; Qwen3-8B abandons \LateWrongOracle\% of decisions with a late change on \orig{}, although the prompt asks it to ignore such changes; Claude keeps all of them. Models react to the announcement of a change to their option and skip the check of when it is in effect, and stronger models skip it less often. Performing the check by rule confirms the diagnosis: on \stress{}, the time filter raises Qwen3-235B from \VAccQwBm\% to \VAccQwBmTime\% (\VPBmTimeBmQwWins--\VPBmTimeBmQwLosses, $p\VPBmTimeBmQw$), while Claude and GPT stay at \VAccClBmTime\% and \VAccGptBmTime\%.

\section{Conclusion}
\label{sec:conclusion}

Freshness is necessary for LLM decision-making and insufficient on its own. A decision is judged when it is carried out, the freshness ceiling bounds what evidence at decision time can guarantee, and the accuracy that remains is set by routing and judgment (Eq.~\ref{eq:decomp}). The controlled tests confirm each prediction: freshness alone leaves changing decisions stale, delivery is sufficient for change and recall follows what the ledger indexes, misleading deliveries cost correct decisions in proportion to weak judgment, and the residual error is mostly temporal. The theory yields three design principles for LLM agents: \emph{record} each decision with its times and what it depends on; \emph{route} news to it until its revision deadline, through an index written when the decision is made; and \emph{revalidate} it by checking when each change is in effect.

\section*{Limitations}

\paragraph{Controlled streams.} The decision streams are synthetic by design: exact control over timing, reference and labels is what lets each prediction be tested in isolation. Real news is longer and noisier, dependencies can be implicit, and relevance can be uncertain, so the magnitudes we report belong to these streams while the predictions concern mechanisms. Testing the theory on logged agent decisions, and in languages beyond English, is the next step.

\paragraph{Scope of the theory.} The theory models a decision as a choice among options with one revision deadline and a binary outcome, and the freshness ceiling assumes a constant hazard rate. Real commitments are staged, partly reversible and graded in utility, and facts change in bursts. The factorisation of Eq.~\ref{eq:decomp} carries over, because information must reach a decision before it can change it, but its terms then depend on time and on the cost of revising.

\paragraph{Reach of the ledger.} Routing recall is bounded by the dependencies the ledger records at decision time, and the ledger treats routed news as trustworthy. Discovering dependencies after the decision and weighing the credibility of sources extend the ledger beyond this paper.

\section*{Ethics Statement}
The decision streams use fictional entities and contain no personal data, and hosted models were queried on this synthetic data only. Systems that revise decisions on a user's behalf can act on false or adversarial news; routing should be paired with source verification and, for consequential decisions, with user confirmation.

\bibliography{references}

@inproceedings{lazaridou2021mind,
  title     = {Mind the Gap: Assessing Temporal Generalization in Neural Language Models},
  author    = {Lazaridou, Angeliki and Kuncoro, Adhiguna and Gribovskaya, Elena and Agrawal, Devang and Li{\v{s}}ka, Adam and Terzi, Tayfun and Gimenez, Mai and de Masson d'Autume, Cyprien and Ko{\v{c}}isk{\'y}, Tom{\'a}{\v{s}} and Ruder, Sebastian and Yogatama, Dani and Cao, Kris and Young, Susannah and Blunsom, Phil},
  booktitle = {Advances in Neural Information Processing Systems},
  volume    = {34},
  year      = {2021}
}

@article{dhingra2022time,
  title   = {Time-Aware Language Models as Temporal Knowledge Bases},
  author  = {Dhingra, Bhuwan and Cole, Jeremy R. and Eisenschlos, Julian Martin and Gillick, Daniel and Eisenstein, Jacob and Cohen, William W.},
  journal = {Transactions of the Association for Computational Linguistics},
  volume  = {10},
  pages   = {257--273},
  year    = {2022}
}

@inproceedings{liska2022streamingqa,
  title     = {{StreamingQA}: A Benchmark for Adaptation to New Knowledge over Time in Question Answering Models},
  author    = {Li{\v{s}}ka, Adam and Ko{\v{c}}isk{\'y}, Tom{\'a}{\v{s}} and Gribovskaya, Elena and Terzi, Tayfun and Sezener, Eren and Agrawal, Devang and de Masson d'Autume, Cyprien and Scholtes, Tim and Zaheer, Manzil and Young, Susannah and Gilsenan-McMahon, Ellen and Austin, Sophia and Blunsom, Phil and Lazaridou, Angeliki},
  booktitle = {Proceedings of the 39th International Conference on Machine Learning},
  series    = {Proceedings of Machine Learning Research},
  volume    = {162},
  year      = {2022}
}

@inproceedings{chen2021timeqa,
  title     = {A Dataset for Answering Time-Sensitive Questions},
  author    = {Chen, Wenhu and Wang, Xinyi and Wang, William Yang},
  booktitle = {Advances in Neural Information Processing Systems (Datasets and Benchmarks Track)},
  year      = {2021}
}

@inproceedings{kasai2023realtime,
  title     = {{RealTime QA}: What's the Answer Right Now?},
  author    = {Kasai, Jungo and Sakaguchi, Keisuke and Takahashi, Yoichi and Le Bras, Ronan and Asai, Akari and Yu, Xinyan and Radev, Dragomir and Smith, Noah A. and Choi, Yejin and Inui, Kentaro},
  booktitle = {Advances in Neural Information Processing Systems (Datasets and Benchmarks Track)},
  year      = {2023}
}

@inproceedings{vu2024freshllms,
  title     = {{FreshLLMs}: Refreshing Large Language Models with Search Engine Augmentation},
  author    = {Vu, Tu and Iyyer, Mohit and Wang, Xuezhi and Constant, Noah and Wei, Jerry and Wei, Jason and Tar, Chris and Sung, Yun-Hsuan and Zhou, Denny and Le, Quoc and Luong, Thang},
  booktitle = {Findings of the Association for Computational Linguistics: ACL 2024},
  year      = {2024}
}

@inproceedings{kim2024carpe,
  title     = {Carpe diem: On the Evaluation of World Knowledge in Lifelong Language Models},
  author    = {Kim, Yujin and Yoon, Jaehong and Ye, Seonghyeon and Bae, Sangmin and Ho, Namgyu and Hwang, Sung Ju and Yun, Se-Young},
  booktitle = {Proceedings of the 2024 Conference of the North American Chapter of the Association for Computational Linguistics: Human Language Technologies (Volume 1: Long Papers)},
  year      = {2024}
}

@inproceedings{fatemi2025testoftime,
  title     = {Test of Time: A Benchmark for Evaluating {LLM}s on Temporal Reasoning},
  author    = {Fatemi, Bahare and Kazemi, Mehran and Tsitsulin, Anton and Malkan, Karishma and Yim, Jinyeong and Palowitch, John and Seo, Sungyong and Halcrow, Jonathan and Perozzi, Bryan},
  booktitle = {International Conference on Learning Representations},
  year      = {2025}
}

@inproceedings{jiang2026benchmarksage,
  title     = {When Benchmarks Age: Temporal Misalignment through Large Language Model Factuality Evaluation},
  author    = {Jiang, Xunyi and Chang, Dingyi and McAuley, Julian and Xu, Xin},
  booktitle = {Proceedings of the 19th Conference of the European Chapter of the Association for Computational Linguistics (Volume 2: Short Papers)},
  pages     = {500--512},
  year      = {2026}
}

@inproceedings{cheng2026temporallyblind,
  title     = {Your {LLM} Agents are Temporally Blind: The Misalignment Between Tool Use Decisions and Human Time Perception},
  author    = {Cheng, Yize and Moakhar, Arshia Soltani and Fan, Chenrui and Hosseini, Parsa and Faghih, Kazem and Sodagar, Zahra and Wang, Wenxiao and Feizi, Soheil},
  booktitle = {Findings of the Association for Computational Linguistics: ACL 2026},
  pages     = {37082--37104},
  year      = {2026}
}

@inproceedings{meng2022rome,
  title     = {Locating and Editing Factual Associations in {GPT}},
  author    = {Meng, Kevin and Bau, David and Andonian, Alex and Belinkov, Yonatan},
  booktitle = {Advances in Neural Information Processing Systems},
  volume    = {35},
  year      = {2022}
}

@inproceedings{zhong2023mquake,
  title     = {{MQuAKE}: Assessing Knowledge Editing in Language Models via Multi-Hop Questions},
  author    = {Zhong, Zexuan and Wu, Zhengxuan and Manning, Christopher D. and Potts, Christopher and Chen, Danqi},
  booktitle = {Proceedings of the 2023 Conference on Empirical Methods in Natural Language Processing},
  year      = {2023}
}

@article{cohen2024ripple,
  title   = {Evaluating the Ripple Effects of Knowledge Editing in Language Models},
  author  = {Cohen, Roi and Biran, Eden and Yoran, Ori and Globerson, Amir and Geva, Mor},
  journal = {Transactions of the Association for Computational Linguistics},
  volume  = {12},
  year    = {2024}
}

@inproceedings{kassner2021beliefbank,
  title     = {{BeliefBank}: Adding Memory to a Pre-Trained Language Model for a Systematic Notion of Belief},
  author    = {Kassner, Nora and Tafjord, Oyvind and Sch{\"u}tze, Hinrich and Clark, Peter},
  booktitle = {Proceedings of the 2021 Conference on Empirical Methods in Natural Language Processing},
  pages     = {8849--8861},
  year      = {2021}
}

@inproceedings{xie2024adaptive,
  title     = {Adaptive Chameleon or Stubborn Sloth: Revealing the Behavior of Large Language Models in Knowledge Conflicts},
  author    = {Xie, Jian and Zhang, Kai and Chen, Jiangjie and Lou, Renze and Su, Yu},
  booktitle = {International Conference on Learning Representations},
  year      = {2024}
}

@inproceedings{xu2024conflicts,
  title     = {Knowledge Conflicts for {LLM}s: A Survey},
  author    = {Xu, Rongwu and Qi, Zehan and Guo, Zhijiang and Wang, Cunxiang and Wang, Hongru and Zhang, Yue and Xu, Wei},
  booktitle = {Proceedings of the 2024 Conference on Empirical Methods in Natural Language Processing},
  year      = {2024}
}

@inproceedings{wilie2024belief,
  title     = {Belief Revision: The Adaptability of Large Language Models Reasoning},
  author    = {Wilie, Bryan and Cahyawijaya, Samuel and Ishii, Etsuko and He, Junxian and Fung, Pascale},
  booktitle = {Proceedings of the 2024 Conference on Empirical Methods in Natural Language Processing},
  pages     = {10480--10496},
  year      = {2024}
}

@article{lyu2026kcbench,
  title   = {{KC-Bench}: A Dynamic Interactive Benchmark for Evaluating Knowledge Conflicts in {LLM} Agents},
  author  = {Lyu, Yaxing and Zhou, Shengjie and Toh, Binbin and Zhu, Pengyu and Li, Lijun},
  journal = {arXiv preprint arXiv:2609.03588},
  year    = {2026}
}

@inproceedings{yao2023react,
  title     = {{ReAct}: Synergizing Reasoning and Acting in Language Models},
  author    = {Yao, Shunyu and Zhao, Jeffrey and Yu, Dian and Du, Nan and Shafran, Izhak and Narasimhan, Karthik and Cao, Yuan},
  booktitle = {International Conference on Learning Representations},
  year      = {2023}
}

@inproceedings{xie2024travelplanner,
  title     = {{TravelPlanner}: A Benchmark for Real-World Planning with Language Agents},
  author    = {Xie, Jian and Zhang, Kai and Chen, Jiangjie and Zhu, Tinghui and Lou, Renze and Tian, Yuandong and Xiao, Yanghua and Su, Yu},
  booktitle = {Proceedings of the 41st International Conference on Machine Learning},
  series    = {Proceedings of Machine Learning Research},
  volume    = {235},
  year      = {2024}
}

@inproceedings{yao2025taubench,
  title     = {$\tau$-bench: A Benchmark for Tool-Agent-User Interaction in Real-World Domains},
  author    = {Yao, Shunyu and Shinn, Noah and Razavi, Pedram and Narasimhan, Karthik},
  booktitle = {International Conference on Learning Representations},
  year      = {2025}
}

@inproceedings{zhang2024timearena,
  title     = {{TimeArena}: Shaping Efficient Multitasking Language Agents in a Time-Aware Simulation},
  author    = {Zhang, Yikai and Yuan, Siyu and Hu, Caiyu and Richardson, Kyle and Xiao, Yanghua and Chen, Jiangjie},
  booktitle = {Proceedings of the 62nd Annual Meeting of the Association for Computational Linguistics (Volume 1: Long Papers)},
  year      = {2024}
}

@inproceedings{lu2025proactive,
  title     = {Proactive Agent: Shifting {LLM} Agents from Reactive Responses to Active Assistance},
  author    = {Lu, Yaxi and Yang, Shenzhi and Qian, Cheng and Chen, Guirong and Luo, Qinyu and Wu, Yesai and Wang, Huadong and Cong, Xin and Zhang, Zhong and Lin, Yankai and Liu, Weiwen and Wang, Yasheng and Liu, Zhiyuan and Liu, Fangming and Sun, Maosong},
  booktitle = {International Conference on Learning Representations},
  year      = {2025}
}

@article{tang2026proagentbench,
  title   = {{ProAgentBench}: Evaluating {LLM} Agents for Proactive Assistance with Real-World Data},
  author  = {Tang, Yuanbo and Tang, Huaze and Cao, Tingyu and Nguyen, Lam and Zhang, Anping and Cao, Xinwen and Liu, Chunkang and Ding, Wenbo and Li, Yang},
  journal = {arXiv preprint arXiv:2602.04482},
  year    = {2026}
}

@article{li2026proevent,
  title   = {{ProEvent}: An Event-centric Benchmark for Proactive Agents},
  author  = {Li, Guanzhen and Pan, Liangming and Wang, Leye},
  journal = {arXiv preprint arXiv:2607.17701},
  year    = {2026}
}

@article{vibelife2026,
  title   = {{VibeLifeBench}: Can Your Life Agent Be Proactive and Persistent in a Living World?},
  author  = {{Xiaohongshu Dots Studio} and {Evolvent AI}},
  journal = {arXiv preprint arXiv:2608.10875},
  year    = {2026}
}

@article{shi2026longterm,
  title   = {Long-term Task-oriented Agent: Proactive Long-term Intent Maintenance in Dynamic Environments},
  author  = {Shi, Qinglong and Wang, Donghai and Zhou, Hantao and Li, Jiguo and Xu, Jun and Gao, Jiuchong and Hao, Jinghua and He, Renqing},
  journal = {arXiv preprint arXiv:2601.09382},
  year    = {2026}
}

@article{zhai2026revisable,
  title   = {Revisable by Design: A Theory of Streaming {LLM} Agent Execution},
  author  = {Zhai, Zhiyuan and Li, Ming and Wang, Xin},
  journal = {arXiv preprint arXiv:2604.23283},
  year    = {2026}
}

@inproceedings{park2023generative,
  title     = {Generative Agents: Interactive Simulacra of Human Behavior},
  author    = {Park, Joon Sung and O'Brien, Joseph C. and Cai, Carrie J. and Morris, Meredith Ringel and Liang, Percy and Bernstein, Michael S.},
  booktitle = {Proceedings of the 36th Annual ACM Symposium on User Interface Software and Technology},
  year      = {2023}
}

@article{packer2023memgpt,
  title   = {{MemGPT}: Towards {LLM}s as Operating Systems},
  author  = {Packer, Charles and Wooders, Sarah and Lin, Kevin and Fang, Vivian and Patil, Shishir G. and Stoica, Ion and Gonzalez, Joseph E.},
  journal = {arXiv preprint arXiv:2310.08560},
  year    = {2023}
}

@inproceedings{wu2025longmemeval,
  title     = {{LongMemEval}: Benchmarking Chat Assistants on Long-Term Interactive Memory},
  author    = {Wu, Di and Wang, Hongwei and Yu, Wenhao and Zhang, Yuwei and Chang, Kai-Wei and Yu, Dong},
  booktitle = {International Conference on Learning Representations},
  year      = {2025}
}

@article{chhikara2025mem0,
  title   = {{Mem0}: Building Production-Ready {AI} Agents with Scalable Long-Term Memory},
  author  = {Chhikara, Prateek and Khant, Dev and Aryan, Saket and Singh, Taranjeet and Yadav, Deshraj},
  journal = {arXiv preprint arXiv:2504.19413},
  year    = {2025}
}

@inproceedings{xu2025amem,
  title     = {{A-Mem}: Agentic Memory for {LLM} Agents},
  author    = {Xu, Wujiang and Liang, Zujie and Mei, Kai and Gao, Hang and Tan, Juntao and Zhang, Yongfeng},
  booktitle = {Advances in Neural Information Processing Systems},
  volume    = {38},
  year      = {2025}
}

@article{rasmussen2025zep,
  title   = {Zep: A Temporal Knowledge Graph Architecture for Agent Memory},
  author  = {Rasmussen, Preston and Paliychuk, Pavlo and Beauvais, Travis and Ryan, Jack and Chalef, Daniel},
  journal = {arXiv preprint arXiv:2501.13956},
  year    = {2025}
}

@article{du2026memory,
  title   = {Memory for Autonomous {LLM} Agents: Mechanisms, Evaluation, and Emerging Frontiers},
  author  = {Du, Pengfei},
  journal = {arXiv preprint arXiv:2603.07670},
  year    = {2026}
}

@inproceedings{lewis2020rag,
  title     = {Retrieval-Augmented Generation for Knowledge-Intensive {NLP} Tasks},
  author    = {Lewis, Patrick and Perez, Ethan and Piktus, Aleksandra and Petroni, Fabio and Karpukhin, Vladimir and Goyal, Naman and K{\"u}ttler, Heinrich and Lewis, Mike and Yih, Wen-tau and Rockt{\"a}schel, Tim and Riedel, Sebastian and Kiela, Douwe},
  booktitle = {Advances in Neural Information Processing Systems},
  volume    = {33},
  year      = {2020}
}

@inproceedings{thakur2021beir,
  title     = {{BEIR}: A Heterogeneous Benchmark for Zero-shot Evaluation of Information Retrieval Models},
  author    = {Thakur, Nandan and Reimers, Nils and R{\"u}ckl{\'e}, Andreas and Srivastava, Abhishek and Gurevych, Iryna},
  booktitle = {Advances in Neural Information Processing Systems (Datasets and Benchmarks Track)},
  year      = {2021}
}

@article{nogueira2019doc2query,
  title   = {Document Expansion by Query Prediction},
  author  = {Nogueira, Rodrigo and Yang, Wei and Lin, Jimmy and Cho, Kyunghyun},
  journal = {arXiv preprint arXiv:1904.08375},
  year    = {2019}
}

@inproceedings{gao2023hyde,
  title     = {Precise Zero-Shot Dense Retrieval without Relevance Labels},
  author    = {Gao, Luyu and Ma, Xueguang and Lin, Jimmy and Callan, Jamie},
  booktitle = {Proceedings of the 61st Annual Meeting of the Association for Computational Linguistics (Volume 1: Long Papers)},
  pages     = {1762--1777},
  year      = {2023}
}

@inproceedings{xiao2024cpack,
  title     = {{C-Pack}: Packed Resources For General {C}hinese Embeddings},
  author    = {Xiao, Shitao and Liu, Zheng and Zhang, Peitian and Muennighoff, Niklas and Lian, Defu and Nie, Jian-Yun},
  booktitle = {Proceedings of the 47th International ACM SIGIR Conference on Research and Development in Information Retrieval},
  year      = {2024}
}

@inproceedings{shi2023distracted,
  title     = {Large Language Models Can Be Easily Distracted by Irrelevant Context},
  author    = {Shi, Freda and Chen, Xinyun and Misra, Kanishka and Scales, Nathan and Dohan, David and Chi, Ed H. and Sch{\"a}rli, Nathanael and Zhou, Denny},
  booktitle = {Proceedings of the 40th International Conference on Machine Learning},
  series    = {Proceedings of Machine Learning Research},
  volume    = {202},
  year      = {2023}
}

@article{liu2024lost,
  title   = {Lost in the Middle: How Language Models Use Long Contexts},
  author  = {Liu, Nelson F. and Lin, Kevin and Hewitt, John and Paranjape, Ashwin and Bevilacqua, Michele and Petroni, Fabio and Liang, Percy},
  journal = {Transactions of the Association for Computational Linguistics},
  volume  = {12},
  pages   = {157--173},
  year    = {2024}
}

@inproceedings{modarressi2025nolima,
  title     = {{NoLiMa}: Long-Context Evaluation Beyond Literal Matching},
  author    = {Modarressi, Ali and Deilamsalehy, Hanieh and Dernoncourt, Franck and Bui, Trung and Rossi, Ryan A. and Yoon, Seunghyun and Sch{\"u}tze, Hinrich},
  booktitle = {Proceedings of the 42nd International Conference on Machine Learning},
  series    = {Proceedings of Machine Learning Research},
  volume    = {267},
  year      = {2025}
}

@article{yang2025qwen3,
  title   = {Qwen3 Technical Report},
  author  = {Yang, An and others},
  journal = {arXiv preprint arXiv:2505.09388},
  year    = {2025}
}

@inproceedings{wolf2020transformers,
  title     = {Transformers: State-of-the-Art Natural Language Processing},
  author    = {Wolf, Thomas and Debut, Lysandre and Sanh, Victor and Chaumond, Julien and Delangue, Clement and Moi, Anthony and Cistac, Pierric and Rault, Tim and Louf, R{\'e}mi and Funtowicz, Morgan and Davison, Joe and Shleifer, Sam and von Platen, Patrick and Ma, Clara and Jernite, Yacine and Plu, Julien and Xu, Canwen and Le Scao, Teven and Gugger, Sylvain and Drame, Mariama and Lhoest, Quentin and Rush, Alexander M.},
  booktitle = {Proceedings of the 2020 Conference on Empirical Methods in Natural Language Processing: System Demonstrations},
  pages     = {38--45},
  year      = {2020}
}

@article{bakkali2026amnesia,
  title   = {Amnesia by Design, Memory By Necessity: Persistent State for Document Intelligence},
  author  = {Bakkali, Souhail and Merimi, Ayoub},
  journal = {arXiv preprint arXiv:2609.32041},
  year    = {2026}
}

@article{jiang2023mistral,
  title   = {Mistral 7{B}},
  author  = {Jiang, Albert Q. and Sablayrolles, Alexandre and Mensch, Arthur and Bamford, Chris and Chaplot, Devendra Singh and de las Casas, Diego and Bressand, Florian and Lengyel, Gianna and Lample, Guillaume and Saulnier, Lucile and Lavaud, L{\'e}lio Renard and Lachaux, Marie-Anne and Stock, Pierre and Le Scao, Teven and Lavril, Thibaut and Wang, Thomas and Lacroix, Timoth{\'e}e and El Sayed, William},
  journal = {arXiv preprint arXiv:2310.06825},
  year    = {2023}
}

@inproceedings{karpukhin2020dpr,
  title     = {Dense Passage Retrieval for Open-Domain Question Answering},
  author    = {Karpukhin, Vladimir and Oguz, Barlas and Min, Sewon and Lewis, Patrick and Wu, Ledell and Edunov, Sergey and Chen, Danqi and Yih, Wen-tau},
  booktitle = {Proceedings of the 2020 Conference on Empirical Methods in Natural Language Processing (EMNLP)},
  pages     = {6769--6781},
  year      = {2020}
}

@article{tian2026swiftmem,
  title   = {{SwiftMem}: Fast Agentic Memory via Query-aware Indexing},
  author  = {Tian, Anxin and Li, Yiming and Li, Xing and Zhen, Hui-Ling and Chen, Lei and Yu, Xianzhi and Dong, Zhenhua and Yuan, Mingxuan},
  journal = {arXiv preprint arXiv:2601.08160},
  year    = {2026}
}

\appendix

\section{Formal Results}
\label{app:proofs}

\paragraph{Proof of Proposition~\ref{prop:ceiling}.}
Let $T$ be the time from the last verification of $f$ until $f$ next changes. A constant hazard rate $\lambda_f$ makes $T$ exponential, $\Pr[T > \tau] = e^{-\lambda_f \tau}$. The fact holds at $t_a$ exactly when $T > t_a - t_v = \Delta_{\mathrm{fresh}} + \Delta_{\mathrm{act}}$, which gives Eq.~\ref{eq:ceiling}. Since $\Delta_{\mathrm{fresh}} \ge 0$, the first factor is at most one, so the probability is at most $e^{-\lambda_f \Delta_{\mathrm{act}}}$, with equality exactly when the evidence is verified at the moment of decision. Fresher evidence moves the probability towards the ceiling; only information that arrives after $t_d$ moves the decision itself. \hfill$\square$

\paragraph{Decisions with several dependencies.}
If the chosen option stays best only while each of the facts $f_1, \dots, f_n$ holds, and the facts change independently, the same argument gives
\begin{equation*}
  \Pr[\text{all } f_k \text{ hold at } t_a] \le \exp\Big(-\Delta_{\mathrm{act}} \sum_{k=1}^{n} \lambda_{f_k}\Big).
\end{equation*}
The ceiling tightens with every dependency, including dependencies the decision text leaves implicit, such as the substation behind a district. This is why the ledger records dependencies with the choice.

\paragraph{Derivation of Eq.~\ref{eq:decomp}.}
A decision is correct when the model keeps it if it should stand and switches it if it should change. Conditioning on the label and, for decisions that should change, on whether a decisive item was delivered, the law of total probability gives
\begin{align*}
  A &= (1-\pi)\Pr[\text{keep} \mid \text{stand}] \\
    &\quad + \pi\rho \Pr[\text{switch} \mid \text{delivered}] \\
    &\quad + \pi(1-\rho)\Pr[\text{switch} \mid \text{missed}],
\end{align*}
which is Eq.~\ref{eq:decomp}. The identity holds for every router and model; the theory's content lies in the predictions about its terms.

\paragraph{Comparing routers.}
Consider two routers used with the same model, with recalls $\rho$ and $\rho'$ and keep rates $K$ and $K'$. If $S$ and $L$ are the same under both, Eq.~\ref{eq:decomp} gives
\begin{equation*}
  A' - A = \pi\,(\rho' - \rho)(S - L) + (1-\pi)(K' - K).
\end{equation*}
A router that reaches more decisive news gains accuracy in proportion to $S - L$, the margin by which the model acts on delivered news over undelivered news, and loses it through the misleading items that lower $K$. Strong judgment makes $S - L$ large and keeps $K$ stable, so recall dominates; weak judgment shrinks $S - L$ and lowers $K$ with every delivery, so exposure dominates. This is prediction P3 in closed form.

\section{Decision Streams}
\label{app:bench}

The generators, the three streams and their SHA-256 hashes are released with the paper; regenerating each stream from its seed reproduces the hash recorded with the model answers. Every label is computed from timestamps and stated facts, and every stream is checked by assertions before use.

\paragraph{\orig{} and \stress{}.}
Decision times fall within the first three days of a six-day stream. Lags are drawn uniformly from 3--12 or 24--72 hours with equal probability, and action times fall in domain windows (dinner 19:00--22:00, trains 06:00--20:00, orders 08:00--20:00, permits on weekdays 09:00--15:30, hikes 07:00--13:00). Decisive items are published inside the revision window and block the option from before $t_a$ until at least an hour after the activity; late items start at least one hour after the activity ends. Two \orig{} notices: ``Notice from Juniper Barrel: closed for a private event from Tue 13 Oct 2026, 14:40 until Wed 14 Oct 2026, 01:10.'' (names the option) and ``City notice: planned power cut in the Riverside district from Wed 14 Oct 2026, 14:40 to Thu 15 Oct 2026, 05:10; restaurants and shops there will close.'' (names the stated dependency). The generator asserts that $t_d < t_r < t_a$, that every decisive item lies in the revision window and overlaps the activity, that no item targets option~2, and that late items never overlap the activity. For \stress{}, it recomputes all labels after placing decoys and paraphrasing notices, and asserts that no gold answer or exclusion changed and that no decoy relates to any decision.

\paragraph{\indirect{}.}
Each domain (walk-in dinners, rail journeys, day hikes, in-store pickups) has 16 bundles of a dependency, its other name, its upstream entity and four facilities. Each bundle serves one decision acting on days 2--3 and one on days 5--6, so the two are never open together; lags are 5--12 or 18--28 hours. A blocking notice for the activity starts between six hours and 15 minutes before it and ends at least 30 minutes after it; a late one starts at least an hour after the activity ends; a past one ends at least 45 minutes before it starts. Late, past and harmless notices use the four ways of reference equally, and blocking, late and past notices share templates, so only their times differ. After placing all notices, the generator recomputes every label and asserts the intended label, exactly two related notices per decision, the balance of the design, that decoys and unrelated notices concern no decision, and that no notice affecting a decision was public before it was made.

\subsection{A Decision from \indirect{}}
\label{app:example}

Decision D0011 is the running example of Figure~\ref{fig:timeline}, reproduced verbatim from the stream. The dossier holds the other name of the district (3), the substation upstream of it (6), four facilities a walking diner does not need (2, 5, 10, 11), and six traps that mention other decisions' entities (1, 4, 7, 8, 9, 12).

\IfFileExists{generated/v3_example.tex}{\begin{small}
\begin{description}[leftmargin=0pt, itemsep=2pt, parsep=0pt, font=\normalfont\itshape]
\item[Times.] Decided Tue 13 Oct 2026, 13:05; revisable until Tue 13 Oct 2026, 19:00; action Tue 13 Oct 2026, 20:30, lasting 2 hours. 30 notices are published in the revision window.
\item[Request.] The user wants to walk in for dinner at 20:30 on Tuesday 13 October 2026 (about 2 hours). They want the best-rated restaurant within a 20-minute walk and will walk there.
\item[Decision-time facts.] ``Lovage Corner: rated 4.6/5; 9-minute walk from the user; in the Bellhaven district; open daily 12:00-02:00; walk-ins welcome.'' ``Red Bistro: rated 4.2/5; 8-minute walk from the user; in the Moorside district; open daily 12:00-02:00; walk-ins welcome.''
\item[Choice.] A. Lovage Corner (alternative: B. Red Bistro).
\item[Dossier.] (1) Tourist boats run from the Pigeon Lofts to the Bellhaven waterfront. (2) The Ironworks museum is in the Bellhaven district. (3) The Bellhaven district is known locally as the Chime Green. (4) The floodlights of the Bellhaven football ground are fed from the Dunsby substation. (5) Tram line T12 stops in the Bellhaven district. (6) The Bellhaven district is supplied with electricity by the Gorran substation. (7) The Bellhaven football club plays its derby against the Southbank district team. (8) Tram line T12 draws its power from the Hensall traction substation. (9) Tram line T12 continues from the Bellhaven district to the Millbrook district. (10) The Bellhaven district hosts the Bread market on weekday mornings. (11) The Bellhaven multi-storey car park serves visitors who drive. (12) Street musicians from the Leafy Triangle often play in the Bellhaven district.
\item[Notices generated with this decision.]
\end{description}
\begin{itemize}[leftmargin=10pt, itemsep=1pt, parsep=0pt]
\item Tue 13 Oct 2026, 14:50, related, harmless: ``City notice: new street lighting installed in the Bellhaven district; no disruption expected.''
\item Tue 13 Oct 2026, 15:05, same-name decoy: ``Road closure: Bellhaven Road in the Ormsby district is shut from Tue 13 Oct 2026, 16:40 to Wed 14 Oct 2026, 01:15.''
\item Tue 13 Oct 2026, 16:40, unrelated: ``Notice from Tidewater Kitchen in the Ormsby district: closed from Tue 13 Oct 2026, 17:10 to Wed 14 Oct 2026, 01:40.''
\item Tue 13 Oct 2026, 18:40, decisive (upstream entity): ``Grid operator: emergency works at the Gorran substation from Tue 13 Oct 2026, 19:50 to Wed 14 Oct 2026, 00:15; every area it supplies will be without power and businesses there will close.''
\end{itemize}
\end{small}

\begin{table}[h]
\centering
\small
\setlength{\tabcolsep}{4pt}
\resizebox{\columnwidth}{!}{%
\begin{tabular}{l rcc ccc}
\toprule
 & \multicolumn{3}{c}{Delivered} & \multicolumn{3}{c}{Answer (gold: B)} \\
\cmidrule(lr){2-4}\cmidrule(lr){5-7}
Router & Items & Decisive & Trap & Claude & Qwen & Ministral \\
\midrule
None & 0 &  &  & A & A & A \\
Name match & 0 &  &  & A & A & A \\
BM25: decision & 6 &  &  & A & A & \textbf{B} \\
BM25: + dossier & 5 & \checkmark & \checkmark & \textbf{B} & \textbf{B} & \textbf{B} \\
Dense: + dossier & 5 &  &  & A & A & A \\
BM25: + watch cond. & 8 & \checkmark & \checkmark & \textbf{B} & \textbf{B} & \textbf{B} \\
Dense: watch cond. & 11 &  &  & A & A & A \\
Oracle: related & 2 & \checkmark &  & \textbf{B} & \textbf{B} & \textbf{B} \\
Oracle: decisive & 1 & \checkmark &  & \textbf{B} & \textbf{B} & \textbf{B} \\
All items & 30 & \checkmark & \checkmark & \textbf{B} & \textbf{B} & A \\
\bottomrule
\end{tabular}}
\caption{Decision D0011 under every router: notices delivered, whether they include the decisive notice and a trap, and each model's answer (bold: correct).}
\label{tab:example}
\end{table}
}{}

Table~\ref{tab:example} follows the decision through every router. The decision text and both dense routers miss the decisive notice, which names only the substation; BM25 over the whole dossier or over the watch conditions reaches it, and every model then switches. Under the decision text, Ministral 8B switches without the decisive notice, prompted by other news: the $L$ term of Eq.~\ref{eq:decomp}.

\section{Prompts}
\label{app:prompts}

Each condition pairs a watch-condition prompt with a re-decision prompt. Every router of a condition uses the same re-decision prompt, so routers differ only in the notices they deliver.

\paragraph{\orig{}.} Both prompts use the system message \emph{``You are an assistant that manages decisions made on behalf of a user. Base your answers only on the information given.''} Watch conditions:
\begin{lstlisting}
A decision has been made for the user and will be acted on later. New information will keep arriving until the revision deadline. Write the decision's watch conditions: the kinds of new information that should make us reconsider it.
Decision time: {t_d}
Evidence at decision time:
[F1] ...
Request: {request}
Options:
A. ...
B. ...
Decision made: {letter}. {option}
The user will act {t_a}. The decision can be changed until {t_r}.
Write 3 to 6 watch conditions. Each must describe a specific event that would make the chosen option unsuitable at the action time. Cover events that affect the chosen option directly, and events that affect it indirectly through the districts, operators, platforms, regions or services it depends on according to the evidence. Name those entities explicitly.
Reply with JSON only, in this form:
{"watch": ["condition 1", "condition 2", "condition 3"]}
\end{lstlisting}
Re-decision:
\begin{lstlisting}
Current time: {t_r}
Evidence at decision time ({t_d}):
[F1] ...
Request: {request}
Options:
A. ...
B. ...
Current decision: {letter}. {option}
New information published since the decision:
[N1] (published {t_w}) ...
The user will act {t_a}. This is the last chance to change the decision. Taking the new information into account, which option should the user go with? Ignore information that is unrelated to these options or that only takes effect after the user's activity is over.
Reply with JSON only: {"answer": "<letter>"}
\end{lstlisting}

\paragraph{\stress{}.} The system message is \emph{``You revise previously made choices using ONLY the facts and notices provided. Respect the action time and notice effective time.''} Claude writes watch conditions with:
\begin{lstlisting}
A decision was made for a later action. Propose 3-6 specific watch conditions that might force revisiting the chosen option.
Use ONLY the decision-time evidence; name the chosen option and its exact dependency entities if stated.
Prioritize notices affecting availability and execution, including indirect disruptions. Avoid generic user-payment, ratings or competing alternatives.
Decision made at: {t_d}
Last revision deadline: {t_r}
Action starts at: {t_a}
Request: {request}
Facts:
[F1] ...
Options:
A. ...
B. ...
Chosen: {letter}. {option}
Reply with JSON only: {"watch":["...","...","..."]}
\end{lstlisting}
Re-decision:
\begin{lstlisting}
Current decision revision time: {t_r}
Decision-time facts ({t_d}):
[F1] ...
Request: {request}
Options:
A. ...
B. ...
Original choice: {letter}. {option}
Newly published notices (shown in time order; may be unrelated or not yet effective):
[N1] (published {t_w}) ...
Action starts: {t_a}. Activity lasts {h} hours.
You must pick the option that will best satisfy the request DURING the action interval, not when the notice was published.
Ignore unrelated notices and changes whose effects start after the activity ends. If the original choice remains best, keep it.
Record only JSON {"answer":"A"} or {"answer":"B"} using the required structured output protocol.
\end{lstlisting}

\paragraph{\indirect{}.} The system message is \emph{``You revise previously made choices using ONLY the facts and notices provided. Respect the action time and when each notice takes effect.''} Claude writes watch conditions with:
\begin{lstlisting}
A decision was made for a user and will be acted on later. New notices will keep arriving until the revision deadline. Write the watch conditions for this decision: the kinds of notice that should make us revisit it.
Decision made at: {t_d}
Revision deadline: {t_r}
Action starts: {t_a}; it lasts {h} hours.
Request: {request}
Decision-time facts:
[F1] ...
Options:
A. ...
B. ...
Chosen: {letter}. {option}
Background facts about the chosen option (from the ledger):
[B1] ...
Write 3 to 8 watch conditions. Each must describe a specific event that would stop the user's activity with the chosen option during the action. Name every entity the activity depends on, including any other name it is known by and anything upstream it relies on. Leave out facilities and entities the user does not need for this activity.
Reply with JSON only: {"watch": ["...", "...", "..."]}
\end{lstlisting}
Re-decision:
\begin{lstlisting}
Current time (last chance to revise): {t_r}
Decision-time facts ({t_d}):
[F1] ...
Request: {request}
Options:
A. ...
B. ...
Original choice: {letter}. {option}
Background facts about the original choice (from the decision ledger):
[B1] ...
Newly published notices (in time order; they may be unrelated or not in effect during the activity):
[N1] (published {t_w}) ...
Action starts: {t_a}. The activity lasts {h} hours.
Pick the option that will best satisfy the request DURING the activity, not when the notice was published.
Ignore notices about other things and changes that are not in effect during the activity. If the original choice remains best, keep it.
Reply with JSON only: {"answer":"A"} or {"answer":"B"}.
\end{lstlisting}

\section{Implementation}
\label{app:impl}

\paragraph{Models and decoding.}
Qwen3-8B runs with thinking disabled through Hugging Face Transformers \citep{wolf2020transformers} with greedy decoding, at most 400 output tokens for watch conditions and 128 for re-decisions. Claude Sonnet 4.6, GPT-6.1 Sol, Qwen3-235B-A22B (2507), Ministral 8B (Ministral 3 8B Instruct) and Mistral 7B (Instruct v0.2) run at temperature 0 with structured output and a plain-JSON fallback, with at most 768 output tokens per answer. Failed or invalid calls are retried and otherwise recorded as missing, never as an answer, and every answer is cached under a hash of the stream, model, router and prompt.

\paragraph{Routing.}
BM25 uses $k_1{=}1.5$ and $b{=}0.75$ over lowercase alphanumeric tokens; dense routing uses BGE-base v1.5 \citep{xiao2024cpack}. Dependency queries use ``\{option\} unavailable cancelled blocked closed or delayed'', ``\{dependency\} outage closure disruption maintenance restriction strike suspended'' and ``\{dependency\} operating hours availability travel safety affects \{option\}''. The time filter parses a start time introduced by \emph{from}, \emph{starting}, \emph{beginning}, \emph{effective} or \emph{start} and drops the decision when that time is not before the end of its activity.

\paragraph{Analysis.}
Every number in the paper is computed by scripts from the per-decision logs, which record for each decision, router and model the notices delivered, whether the decisive notice and traps were among them, and the answer. The scripts and logs are released with the paper.

\section{Additional Results}
\label{app:more}

\paragraph{Where the errors are.} Table~\ref{tab:v3types} breaks down \indirect{} accuracy by how the decisive notice refers to the option and by the related news a decision received. Routing sets which switches are possible: the decision text reaches the option and its stated dependency, and the dossier and watch conditions extend the reach to other names and upstream entities. Judgment sets the rest: decisions with a change that is already over or starts after the activity are the ones most often abandoned.

\IfFileExists{generated/v3_types.tex}{\begin{table}[t]
\centering
\small
\setlength{\tabcolsep}{3pt}
\resizebox{\columnwidth}{!}{%
\begin{tabular}{l rrrrrr}
\toprule
 & Name & BM25 & +Doss. & +Watch & Oracle & All \\
\midrule
\multicolumn{7}{l}{\emph{Needed switches made, by how the notice refers to the option}} \\
\quad option name & 100.0 & 100.0 & 100.0 & 100.0 & 100.0 & 100.0 \\
\quad dependency & 8.3 & 100.0 & 100.0 & 97.9 & 100.0 & 97.9 \\
\quad other name & 6.2 & 54.2 & 95.8 & 97.9 & 97.9 & 93.8 \\
\quad upstream entity & 4.2 & 35.4 & 93.8 & 75.0 & 100.0 & 89.6 \\
\multicolumn{7}{l}{\emph{Decisions kept, by the related news they received}} \\
\quad starts after & 81.2 & 52.1 & 37.5 & 43.8 & 43.8 & 39.6 \\
\quad already over & 77.1 & 37.5 & 27.1 & 33.3 & 33.3 & 27.1 \\
\quad harmless & 91.7 & 68.8 & 52.1 & 70.8 & 72.9 & 52.1 \\
\quad not needed & 89.6 & 70.8 & 52.1 & 60.4 & 62.5 & 43.8 \\
\bottomrule
\end{tabular}}
\caption{Needed switches made and decisions kept on \textsc{Indirect}, by how the decisive notice refers to the option and by the related news a decision received (\%, mean over the three models with complete answers; 16 decisions per row and model). Routing sets which switches are possible; judgment sets the rest.}
\label{tab:v3types}
\end{table}
}{}

\paragraph{Statistical tests.} Table~\ref{tab:v3tests} lists the exact McNemar tests behind the \indirect{} comparisons in Section~\ref{sec:results}.

\IfFileExists{generated/v3_tests.tex}{\begin{table*}[t]
\centering
\small
\resizebox{\textwidth}{!}{%
\begin{tabular}{ll rr rr rr rr}
\toprule
 & & \multicolumn{2}{c}{Claude} & \multicolumn{2}{c}{Qwen-235B} & \multicolumn{2}{c}{Ministral 8B} & \multicolumn{2}{c}{Pooled} \\
\cmidrule(lr){3-4}\cmidrule(lr){5-6}\cmidrule(lr){7-8}\cmidrule(lr){9-10}
Router A & Router B & W--L & $p$ & W--L & $p$ & W--L & $p$ & W--L & $p$ \\
\midrule
BM25: decision & Name match & 26--5 & ${<}.001$ & 30--26 & .689 & 30--26 & .689 & 86--57 & .019 \\
BM25: + dossier & BM25: decision & 22--4 & ${<}.001$ & 17--15 & .860 & 12--13 & 1.000 & 51--32 & .048 \\
BM25: + watch cond. & BM25: decision & 21--5 & .002 & 16--11 & .442 & 15--7 & .134 & 52--23 & .001 \\
BM25: + watch cond. & BM25: + dossier & 5--7 & .774 & 9--6 & .607 & 12--3 & .035 & 26--16 & .164 \\
Dense: watch cond. & Dense: + dossier & 11--13 & .839 & 12--16 & .572 & 10--9 & 1.000 & 33--38 & .635 \\
All items & BM25: + dossier & 6--8 & .791 & 2--9 & .065 & 10--8 & .815 & 18--25 & .360 \\
All items & BM25: + watch cond. & 6--6 & 1.000 & 6--16 & .052 & 8--15 & .210 & 20--37 & .033 \\
Oracle: related & BM25: + dossier & 9--1 & .021 & 11--0 & ${<}.001$ & 11--5 & .210 & 31--6 & ${<}.001$ \\
Oracle: decisive & Oracle: related & 1--0 & 1.000 & 38--0 & ${<}.001$ & 52--0 & ${<}.001$ & 91--0 & ${<}.001$ \\
\bottomrule
\end{tabular}}
\caption{Exact McNemar tests on \textsc{Indirect}. W--L: decisions router A gets right and router B gets wrong, and the reverse; 128 decisions per model, pooled over the three models with complete answers. $p$-values are two-sided and uncorrected.}
\label{tab:v3tests}
\end{table*}
}{}

\paragraph{Every router on \orig{}.}\label{app:orig}
Table~\ref{tab:main} reports every router with Qwen3-8B on \orig{}, and Table~\ref{tab:needless} breaks down its needless switches by the related news a decision received. With oracle routing, Qwen3-8B makes every needed switch and keeps \KeepOracle\% of the decisions that should stand. BM25 over the decision text, scored on \orig{} for routing only, delivers \VOneBmRec\% of decisive pairs.

\IfFileExists{generated/main_results.tex}{\begin{table*}[t]
\centering
\small
\setlength{\tabcolsep}{5pt}
\resizebox{\textwidth}{!}{%
\begin{tabular}{l ccc c ccc c c}
\toprule
 & \multicolumn{3}{c}{Routing (\%)} & & \multicolumn{3}{c}{Final decisions (\%)} & & \\
\cmidrule(lr){2-4}\cmidrule(lr){6-8}
Router & Recall & Implicit & Precision & Accuracy & Switch: named & Switch: dependency & Keep & Tokens (k) & $p$ \\
\midrule
\multicolumn{10}{l}{\textit{Qwen/Qwen3-8B}} \\
None (freshness at $t_d$) & 0.0 & 0.0 & -- & 45.5 & 0.0 & 0.0 & 100.0 & 0.0 & <0.001 \\
Name match & 39.7 & 0.0 & 100.0 & 70.0 & 100.0 & 18.2 & 88.0 & 23.9 & 0.851 \\
Dense similarity & 83.8 & 75.6 & 14.3 & \textbf{71.8} & 92.0 & 63.6 & 66.0 & 102.4 & -- \\
Watch conditions & 77.9 & 65.9 & 14.2 & 62.7 & 80.0 & 57.6 & 56.0 & 146.9 & 0.076 \\
Oracle: related & 100.0 & 100.0 & 100.0 & 82.7 & 100.0 & 100.0 & 62.0 & 44.2 & 0.050 \\
All items & 100.0 & 100.0 & 1.6 & 55.5 & 60.0 & 33.3 & 68.0 & 803.7 & 0.003 \\
\bottomrule
\end{tabular}}
\caption{Qwen3-8B on \orig{}. \emph{Recall}: share of decision-changing items delivered to the decision they change; \emph{Implicit}: the same for items that reach the option only through a dependency; \emph{Precision}: share of delivered items that concern the chosen option. \emph{Switch: named / dependency}: share of decisions that should switch and did, when the evidence names the option or reaches it only through a dependency; \emph{Keep}: share of decisions that should stand and did. \emph{Tokens}: prompt tokens spent on routing and re-deciding, including writing watch conditions. $p$: exact McNemar test on per-decision correctness against the most accurate practical router (bold; oracle excluded). Routing recall and precision for None, Name match, Oracle and All follow from the stream and are identical across models.}
\label{tab:main}
\end{table*}
}{}
\IfFileExists{generated/needless.tex}{\begin{table}[t]
\centering
\small
\setlength{\tabcolsep}{4pt}
\begin{tabular}{l r ccccc}
\toprule
Related news & $n$ & Oracle & Name & Sim. & Watch & All \\
\midrule
Late change & 18 & 94 & 28 & 39 & 67 & 33 \\
Harmless only & 19 & 11 & 5 & 37 & 37 & 32 \\
None & 13 & 0 & 0 & 23 & 23 & 31 \\
\bottomrule
\end{tabular}
\caption{Needless switches (\%) on decisions that should stand, by the related news published in their revision window. \emph{Late change}: a change to the option or its dependency that starts only after the activity ends; \emph{Harmless only}: news about the option or its dependency that does not affect it. Name: name match; Sim.: similarity to the decision; Watch: watch conditions.}
\label{tab:needless}
\end{table}
}{}

\paragraph{Routing with other $K$.} Table~\ref{tab:v3ksweep} shows \indirect{} routing as the number $K$ of decisions per notice varies. With $K{=}1$, the whole dossier delivers \WKOneBmDoss\% of decisive notices and watch conditions \WKOneWatchBm\%.

\IfFileExists{generated/v3_ksweep.tex}{\begin{table}[h]
\centering
\small
\setlength{\tabcolsep}{3.5pt}
\resizebox{\columnwidth}{!}{%
\begin{tabular}{l rr rr rr}
\toprule
 & \multicolumn{2}{c}{$K{=}1$} & \multicolumn{2}{c}{$K{=}3$} & \multicolumn{2}{c}{$K{=}5$} \\
\cmidrule(lr){2-3}\cmidrule(lr){4-5}\cmidrule(lr){6-7}
Router & Recall & Traps & Recall & Traps & Recall & Traps \\
\midrule
BM25: decision & 51.6 & 4 & 62.5 & 17 & 68.8 & 35 \\
BM25: + dossier & 85.9 & 21 & 100.0 & 89 & 100.0 & 96 \\
Dense: + dossier & 70.3 & 11 & 85.9 & 59 & 90.6 & 74 \\
BM25: + watch cond. & 71.9 & 5 & 92.2 & 20 & 96.9 & 38 \\
Dense: watch cond. & 75.0 & 4 & 84.4 & 27 & 90.6 & 53 \\
\bottomrule
\end{tabular}}
\caption{\textsc{Indirect} routing as the number $K$ of decisions each notice is sent to varies: decisive pairs delivered (\%) and lexical traps delivered.}
\label{tab:v3ksweep}
\end{table}
}{}

\paragraph{A second \stress{} sample.} Table~\ref{tab:second} shows Claude Sonnet 4.6 on a second sample of 73 decisions, where routing again lifts accuracy from \VRobAccNone\% without routing to \VRobAccBm\% with BM25, against \VRobAccOrRel\% for the related oracle.

\IfFileExists{generated/v2_second.tex}{\begin{table}[h]
\centering
\small
\begin{tabular}{l rrr}
\toprule
Router & Accuracy & Keep & Needless \\
\midrule
None & 61.6 & 100.0 & 0 \\
Name match & 75.3 & 91.1 & 4 \\
BM25: decision & 89.0 & 91.1 & 4 \\
BM25: + time filter & 91.8 & 93.3 & 3 \\
Dense: decision & 91.8 & 93.3 & 3 \\
Dense: watch cond. & 89.0 & 86.7 & 6 \\
Dependency queries & 90.4 & 88.9 & 5 \\
Oracle: related & 94.5 & 93.3 & 3 \\
Oracle: decisive & 100.0 & 100.0 & 0 \\
\bottomrule
\end{tabular}
\caption{Claude Sonnet 4.6 on a second \textsc{Stress} sample (73 decisions). Keep: accuracy on decisions that should stand (\%); Needless: switches on those decisions. The table lists the routers run on this sample.}
\label{tab:second}
\end{table}
}{}

\paragraph{Mistral 7B on \indirect{}.} Mistral 7B returned a valid answer for 74--95\% of the prompts, depending on the router, and mostly kept its original choice: given only decisive notices, it switched on 48.9\% of the decisions it answered. The comparisons in Section~\ref{sec:results} use the three models with complete answers.

\end{document}